\documentclass[letterpaper, 10 pt, conference]{ieeeconf}  

\IEEEoverridecommandlockouts                              

\usepackage{graphicx}
\usepackage{color,soul}
\usepackage{hyperref}
\usepackage{mwe}
\usepackage{subcaption}
\usepackage{amsmath}
\usepackage{tabularx} 
\usepackage{multirow}

\newcolumntype{L}{>{\centering\arraybackslash}m{3cm}}
\usepackage{float}
\usepackage[thinc]{esdiff}
\usepackage{graphics} 
\usepackage[table]{xcolor}
\usepackage{caption}
\usepackage{graphicx}

\usepackage{graphics} 
\usepackage{epsfig} 
\usepackage{mathptmx} 
\usepackage{times} 
\usepackage{amsmath} 
\usepackage{amssymb}  
\usepackage[noadjust]{cite}
\usepackage{booktabs}
\usepackage{multirow}
\usepackage{algorithmic}
\usepackage{ragged2e}
\usepackage{array}

\newcolumntype{P}[1]{>{\RaggedRight\arraybackslash}p{#1}}
\newcolumntype{M}[1]{>{\centering\arraybackslash}m{#1}}

\usepackage{microtype}
\renewcommand{\baselinestretch}{1.00}

\usepackage[table]{xcolor}

\title{\Large \bf Field Insights for Portable Vine Robots in Urban Search and Rescue}
\author{{\scalebox{0.96}{Ciera McFarland$^{1*}$, Ankush Dhawan$^{2,3*}$, Riya Kumari$^{2}$, Chad Council$^{2}$, Margaret Coad$^{1\dag}$, and Nathaniel Hanson$^{2\dag}$}
\thanks{$^{*}$Equal contribution $^{\dag}$Equal contribution}
\thanks{Correspondence: {\tt\footnotesize nathaniel.hanson@ll.mit.edu}}}%
\thanks{$^{1}$University of Notre Dame, Notre Dame, Indiana, USA}
\thanks{$^{2}$Lincoln Laboratory, Massachusetts Institute of Technology, Lexington, Massachusetts, USA}
\thanks{$^{3}$Stanford University, Stanford, California, USA}
\thanks{DISTRIBUTION STATEMENT A. Approved for public release. Distribution is unlimited.
 This material is based upon work supported by the Department of the Air Force under Air Force Contract No. FA8702-15-D-0001. Any opinions, findings, conclusions or recommendations expressed in this material are those of the author(s) and do not necessarily reflect the views of the Department of the Air Force.}
}

\begin{document}

\maketitle
\thispagestyle{empty}
\pagestyle{empty}

\begin{abstract}

Soft, growing vine robots are well-suited for exploring cluttered, unknown environments, and are theorized to be performant during structural collapse incidents caused by earthquakes, fires, explosions, and material flaws. These vine robots grow from the tip, enabling them to navigate rubble-filled passageways easily. State-of-the-art vine robots have been tested in archaeological and other field settings, but their translational capabilities to urban search and rescue (USAR) are not well understood. To this end, we present a set of experiments designed to test the limits of a vine robot system, the Soft Pathfinding Robotic Observation Unit (SPROUT), operating in an engineered collapsed structure. Our testing is driven by a taxonomy of difficulty derived from the challenges USAR crews face navigating void spaces and their associated hazards. Initial experiments explore the viability of the vine robot form factor, both ideal and implemented, as well as the control and sensorization of the system. A secondary set of experiments applies domain-specific design improvements to increase the portability and reliability of the system. SPROUT can grow through tight apertures, around corners, and into void spaces, but requires additional development in sensorization to improve control and situational awareness. 


\end{abstract}

\section{INTRODUCTION}

In the event of a building collapse, the reduction of a structure to an unstructured debris pile may create void spaces -- internal, subsurface pockets where human survivors may be located \cite{firehouse_collapse_2018}. Collapses present a myriad of electrical, chemical, and mechanical hazards to rescue crews resulting from exposed wires, gas leaks, sharp objects, and unstable debris among many other dangers \cite{osha_structural_collapse}. Amidst these hazards, organized Urban Search and Rescue (USAR) teams, including canines and their handlers, structural engineers, and technical search specialists jointly assess the debris to locate survivors and identify the safest entry points in the structure using specialized search cameras to probe just beneath the surface.

USAR teams have long employed rescue robots of varying morphologies to augment response efforts and accelerate response times \cite{Murphy2016}. Drones, unmanned ground vehicles \cite{Surmann2022lessons, murphy2017disaster} and legged robots \cite{Yoshiike2017Legged, osborne2023robotdog, hoeller2024anymal} enable above-ground monitoring and mapping of collapsed sites. Snake-like robots \cite{Whitman2018uSnake, Ambe2016Kumamoto, Fujikawa2019asc} have explored subsurface areas in collapsed buildings. Nonetheless, these established forms tend to be both expensive and rigid, which limits their adaptability to access small voids. In these environments where secondary collapse and robot loss are ever-present risks, there is an unmet need for a low-cost, capable platform to explore the void space for survivors of these tragedies and viable paths to reach them.

\begin{figure}[!tbp]
\centering
\includegraphics[width = \columnwidth]{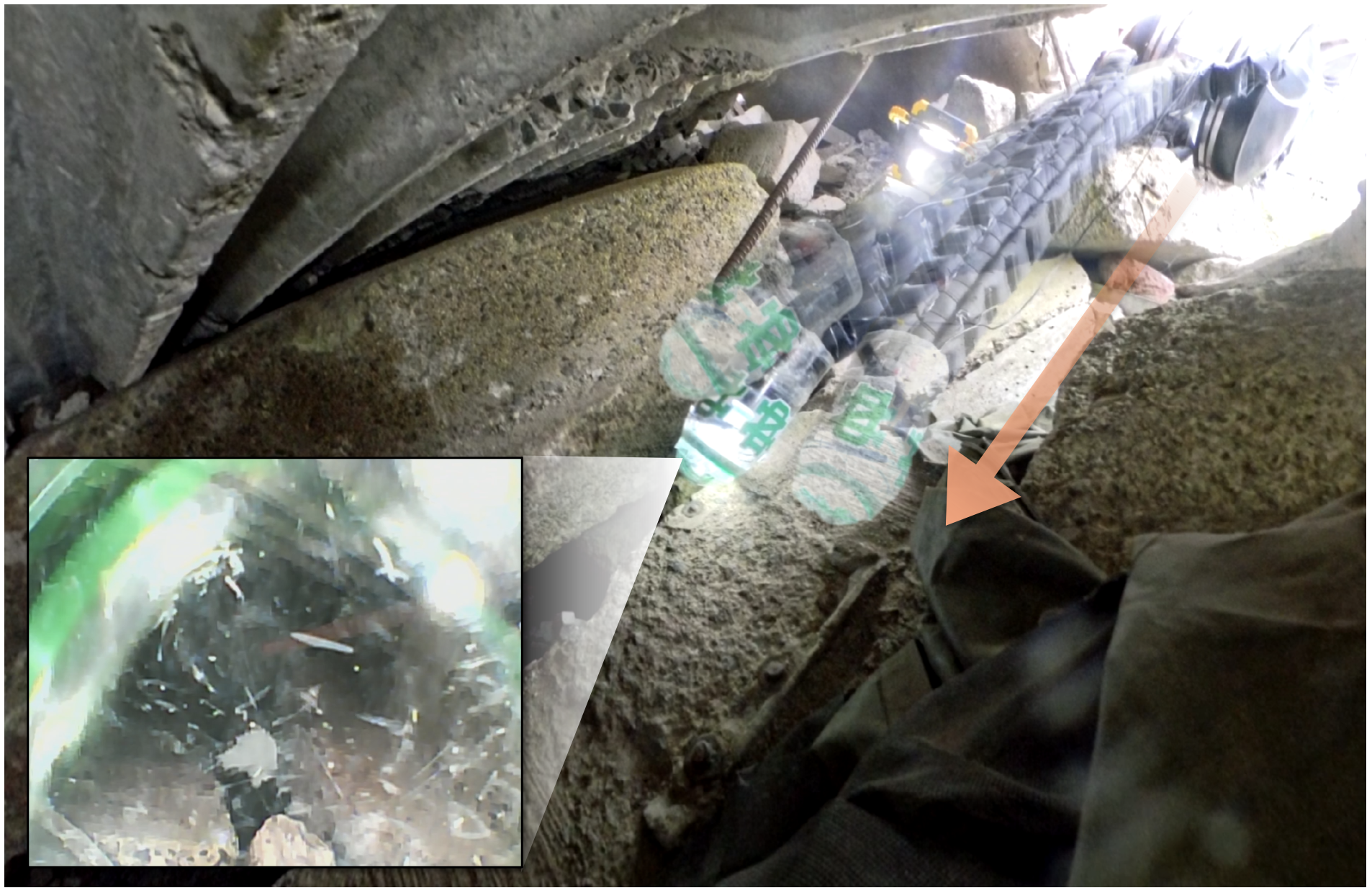}
\caption{\textbf{External view of SPROUT growing into the void space inside a mock collapsed structure.} Orange arrow shows the direction of growth from outside the void. Transparent images depict robot pointing its sensor head in multiple directions. (Inset) \textbf{Egocentric view of space using tip-mounted camera.}}
\label{fig:gs}
\vspace{-2.0em}
\end{figure}

To address the challenges of this space, we consider vine robots: soft, growing structures that evert from their tips \cite{hawkes2017soft}. They are driven by air pressure, squeeze through small holes, move with minimal disturbance to the environment \cite{blumenschein2020design}, and are steered with pneumatic actuators \cite{greer2017series}. In sum, they travel along deliberate, complex paths through 3D space from user directions \cite{el2018development} and carry sensors at their tips \cite{blumenschein2020design}. 


In this paper, we adapt our robot previously used successfully in archaeological field studies \cite{coad2020vine} for integration with USAR teams. Our system, known as SPROUT -- the Soft Pathfinding Robotic Observation Unit, is evaluated on the training site for Massachusetts Task Force~1 (MA-TF1). This study contextualizes recent developments in vine robots and their adaptability to the needs of the USAR community.


The contributions of this work are:
\begin{itemize}
    \item A taxonomy of difficulty to categorize challenges presented by collapsed structure void spaces.
    \item Analysis of state-of-the-art vine robot technology in a realistic debris pile with multiple points of entry.
    \item System modifications of a vine robot for increased portability and on-pile operations.
\end{itemize}

\section{RELATED WORK}
Various robots have been deployed in the field and in search-and-rescue contexts. Here, we present a short review of serpentine rescue robots, which are similar to vine robots in form, but not in locomotive mechanism. We also overview relevant developments from the vine robotics literature that have informed our current system design.

\subsection{Serpentine Rescue Robots}

Snake-like robots demonstrate the ability to explore narrow areas under debris. An 8~m long flexible continuum robot with sensors, named ``Active Scope Camera" (ASC), has been used in several disaster sites, such as to investigate collapsed houses in the 2016 Kumamoto earthquake~\cite{Ambe2016Kumamoto} and to investigate a collapsed building with 8~m deep vertical shafts~\cite{Fukuda2014verticalASC}. It has a ciliary vibration drive that propels the robot body forward. An improved ASC design~\cite{Fujikawa2019asc} included an active air jet nozzle on the head to allow for better mobility, operability, durability, and speed, as tested in a mock disaster environment. However, rescue workers expressed concerns regarding the dust and sound generated by the air injection, as well as a preference for a smaller and lighter system. Vine robots are beneficial in this regard as their lightweight bodies allow them to support their weight with relative ease in order to access spaces. 

 Other snake-like robot designs, such as U-snake \cite{Whitman2018uSnake}, were used to assist in the 2017 Mexico City Earthquake. The body of the robot is made up of sixteen identical modules, with each joint axis offset by 90 degrees from the previous. Different types of terrains are traversed by executing different gaits. The different snake gaits allowed for movement over obstacles and the ability to control the head camera to look around corners, providing visual input exceeding the capabilities of conventional search cameras. However, this snake robot was tethered, and it operated best with a tether manager acting in addition to the teleoperator to keep the tether from becoming caught on the environment. One advantage of vine robots is that their body acts as their tether, generally eliminating the need to have a separate operator managing that aspect. The nature of the tether and teleoperation tasks for the snake robot required the operators to work at a distance from each other, which led to communication difficulties. 

\subsection{Vine Robots in the Field}
There are two main published examples of vine robots tested in outdoor field environments. One vine robot system was deployed both at a soft robotics competition course that modeled a disaster and at an archaeological site \cite{coad2020vine}. This vine robot grows from a base, has three pneumatic actuators running along its entire length, and is teleoperated by a joystick. The competition featured a sand pit, square aperture, stairs, and a group of unstable cylinders that were meant to remain upright after the robot passed through the area. The vine robot was able to successfully clear all four obstacles, demonstrating its usefulness in navigating challenging spaces without damaging the environment. The robot was then deployed in narrow tunnels at an archaeological site that humans could not enter. In this setting, the robot grew over rocks and up a vertical shaft, and completed a \mbox{90$^{\circ}$} turn all while carrying a camera at its tip. 


Additionally, a vine robot with a different design has been deployed in an outdoor rubble pile used as a  USAR training site~\cite{der2021roboa}. This robot also grows from a compact base and is teleoperated by a joystick, but the robot is steered at its tip rather than along its entire body. The robot was tasked with finding a person ``hidden" in the rubble using only its tip-mounted camera for visual feedback on the robot's progress. It successfully moved over debris and completed a \mbox{90$^{\circ}$} turn from an entry point located at the base of the pile. This robot was well-received by the Swiss Rescue Troop members who supervised the test, demonstrating confidence in soft robots to provide assistance in these types of environments. However, the heavy tip mount prevented the robot from exploring the vertical complexity of the pile's void space; moreover, the relatively bulky base unit prevents easy transport onto the pile for insertion into void spaces. Thus, there is still an unmet need in the literature, but more importantly for practitioners, for a low-cost, field portable robot to explore collapsed structures.

\section{TECHNICAL APPROACH}

\begin{figure}[tb]
    \vspace{0.5em}
    \centering
    \includegraphics[width=\linewidth]{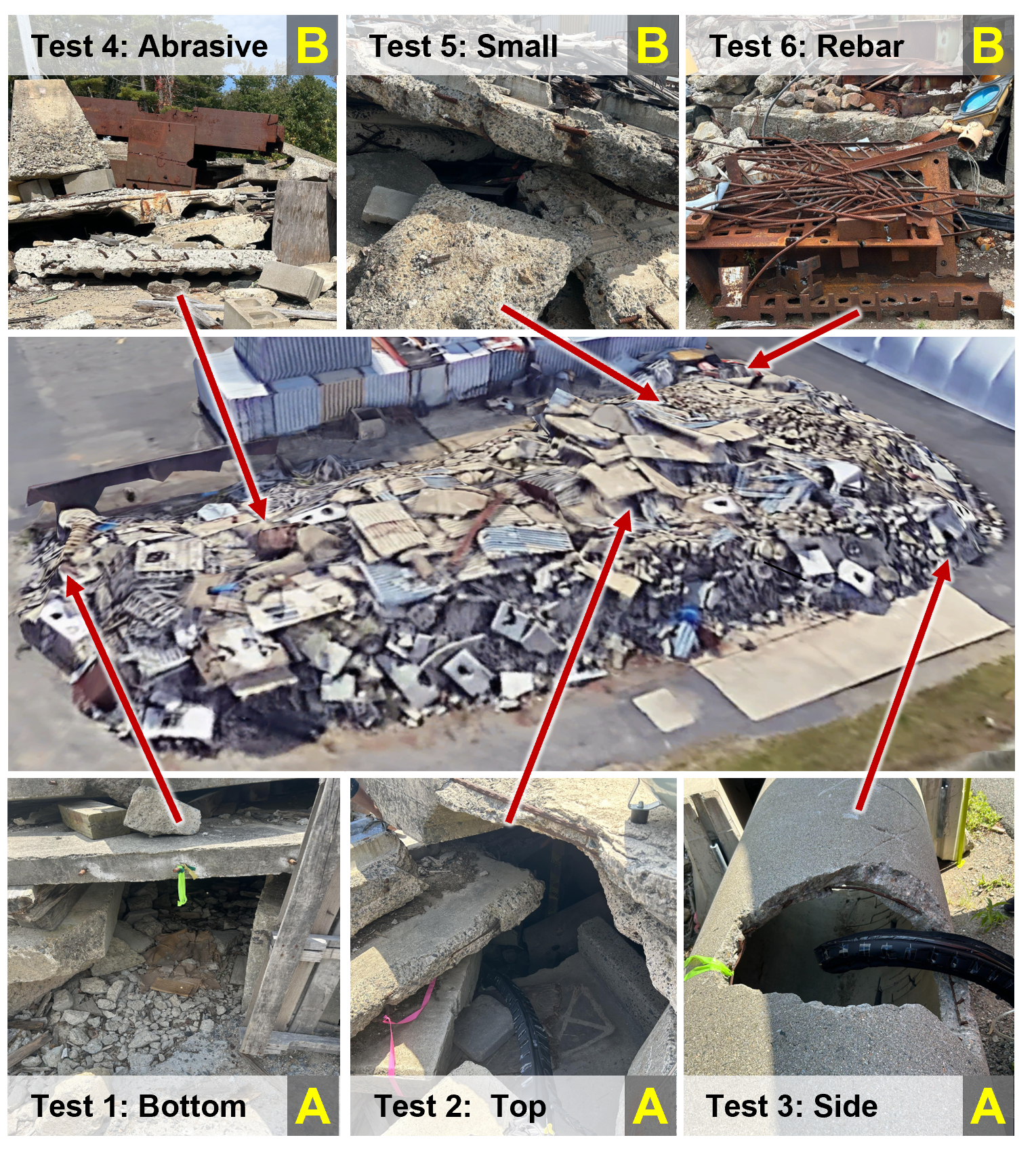}
    \caption{\textbf{Test locations on Massachusetts Task Force~1 rubble pile.} We tested SPROUT at six locations on the mock collapsed structure rubble pile in Beverly, MA, each corresponding to different levels on the taxonomy of difficulty from Table~\ref{table:tod}. The top and bottom rows of images present close-up views of the trial locations from the second field study (B) and the initial field study (A). Middle image is a birds-eye view of the rubble pile. Map data: Google 2024.}
    \label{fig:test_locs}
    \vspace{-1.5em}
\end{figure}

Our research into the feasibility of vine robots in USAR operations was structured in a two-phase approach. During the initial phase, we brought an existing vine robot setup, largely unchanged from previous field experiments \cite{coad2020vine}, to the Massachusetts Task Force~1's (MA-TF1) training site in Beverly, MA. This site is a man-made engineered collapsed structure rubble pile with many entry points of varying difficulty to traverse. Under consultation from technical search specialists, we deployed the system in multiple different locations on the pile, shown in Fig.~\ref{fig:test_locs}. To guide our testing, we developed a \textit{Taxonomy of Difficulty} to categorize the operational hazards in collapsed structure response.

After noting the challenges and opportunities of the domain, we implemented several improvements to SPROUT to accommodate the rugged environment of the MA-TF1 site. These changes between the visits improved SPROUT's portability and durability during field deployments. From both field studies, we document system performance --- both positive and negative --- that highlights trends across the variety of surface types and spaces found in a rubble pile.

\subsection{Taxonomy of Difficulty}

\begin{table}[tb]
\centering
\vspace{1ex}
\begin{tabular}{P{0.5cm} P{2.9cm} P{1.25cm} P{1.4cm} P{0.775cm}}
\toprule
\textbf{Level} & \textbf{Description} & \textbf{Human Challenge} & \textbf{Vine Robot Challenge} & \textbf{Test} \\ \midrule
1 & Completely straight path & Easy & Easy & 1, 2 \\
 2 & Horizontal path between \par entry and destination & Easy & Easy & 1, 5, 6 \\
3 & Horizontal path becomes vertical and down & Medium & Easy & 3, 4 \\
4 & Vertical entry point & Hard & Easy & 2, 3 \\
5 & Entry point too small for a human & Impossible & Easy & 3, 5, 6 \\
6 & Loose debris or thin obstacles (rebar/cables) & Medium & Easy & 4, 6 \\
7 & Pooled water or slurry & Medium & Medium & - \\
8 & Utility hazards (live electricity, gas leak, etc.) & Hard & Unknown & - \\
\bottomrule
\end{tabular}
\caption{\textbf{Taxonomy of Difficulty for void space entry for humans and vine robots.} Test locations, highlighted in Fig.~\ref{fig:test_locs}, are mapped to levels on the taxonomy of difficulty based on their characteristics.}
\label{table:tod}
\end{table}

While failure modes for robots in disaster scenarios have structured taxonomies \cite{carlson2005ugvs}, there is not a comparable method used by practitioners to assess  personal risk in entering void spaces. We consulted USAR experts in the Federal Emergency Management Agency (FEMA), including structural collapse technicians and specialists and task force leadership, to develop a categorical understanding of hazards in void spaces. We then leveraged our prior robotics work to compare the level of human challenge to that faced by a vine robot under the same conditions. We use this resulting \textit{Taxonomy of Difficulty} to categorize entry into void spaces in structural collapse, ordered in levels of increasing difficulty and physical hazards as rated by our expert human responders. Table~\ref{table:tod} shows the eight levels of the taxonomy of difficulty, including the difficulty rating for a human, the expected difficulty for a vine robot based on previous literature, and where the test locations for both our field studies map onto the taxonomy. One key observation is that when the physical spaces become increasingly difficult for a human to enter, an ideal vine robot should be able to overcome the same obstacles with better ease. With respect to levels 7 and 8 of the taxonomy, vine robots have also been demonstrated to navigate over water~\cite{hawkes2017soft}, but have not been tested in the presence of utility hazards. The MA-TF1 did not contain these features, but they are included for completeness.

\begin{figure*}[tb]
\vspace{3ex}
\centering
\includegraphics[width = \linewidth]{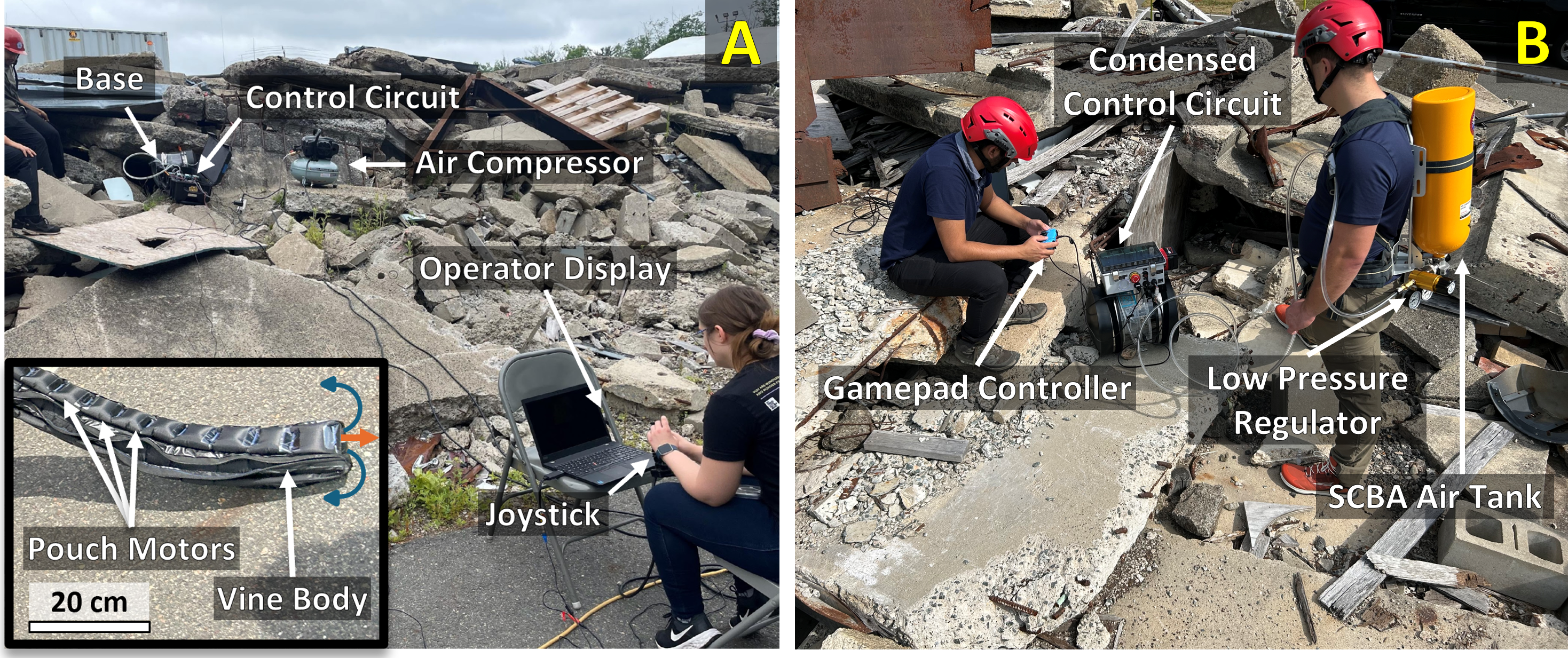}
\caption{\textbf{Experimental setup for field studies.} (A) The SPROUT system that we tested in the initial field study consists of a base containing the spooled vine body, a portable air compressor, and control electronics in separate cases. The operator controls the system with a joystick while viewing the tip-mounted camera stream on a screen, allowing safe operation if the entry point into the pile is unstable or hazardous. (Inset) View of SPROUT with tip camera removed showing the principle of growth by eversion. (B) Improved SPROUT system incorporating feedback from urban search and rescue team members and insights from the first study. System condenses control electronics into ingress-protected container which mounts to the top of the pressurized vine base. A Self Contained Breathing Apparatus (SCBA) tank provides compressed air to grow and steer the robot.}

\label{fig:parts_image}
\vspace{-3ex}
\end{figure*}

\section{INITIAL FIELD STUDY}
\subsection{Initial System Overview}
The main component of the SPROUT system, Fig.~\ref{fig:parts_image}A, is an inflatable vine body made of a tube of airtight fabric. The proximal end of the vine body is fixed to an outlet tube on the base, first introduced in \cite{coad2020vine}, and the distal end is attached to a cord the length of the body. Before growth, the cord and vine body are wrapped around a spool inside the base. To grow, the base and the vine body are pressurized with air, and the spool's motor/encoder is commanded to unroll, allowing the robot to evert and extend from its tip. 

To steer SPROUT, three series pouch motor actuators~\cite{coad2020vine} are spaced around the circumference of the vine body and attached along its length. When each actuator is pressurized with air, it shortens, causing the vine body to curve towards that actuator. The actuators can be inflated in combination to steer in any direction.

SPROUT carries a camera using a rounded camera mount that loosely fits over and gets pushed along by the vine body tip as it extends. The camera setup includes LED lights wired to the camera's power supply. Power and signal for the camera run over a cable that moves freely outside the robot body. 

SPROUT is teleoperated using a joystick, originally introduced in \cite{el2018development}, that allows a human operator to send desired growth speed and curvature commands over a wire to the robot while sitting a safe distance away. The human operator views the image from the camera at the robot tip via a display.

The pneumatic components of SPROUT are supplied by an air compressor, which, along with the electrical components, is powered by a 120V AC power line.

\subsection{Initial System Fabrication}
The SPROUT body for the initial field study was constructed from 40 denier TPU-coated ripstop nylon (extremtextil, Dresden, Germany), with a thickness of 0.031~mm, and a calculated density of 2200~kg/m\(^3\). The parameters for our robot are shown in Table~\ref{table:robot parameters}, along with those of the robot we used in the second field study. All tubes are sealed with an impulse sealer (E-MFS-450, Technopack, Sunrise, FL). The seals are sewn over to increase their strength. The actuators are attached to the robot body using double-sided tape (MD-9000, Marker-Tape, Canon City, CO).  

\begin{table}[b]
\vspace{-3ex}
\begin{center}
\begin{tabular}{ c|c|c}
\hline
\textbf{Measurement} & \textbf{Initial Field Study} & \textbf{Second Field Study} \\
\hline
Body Diameter  & 12.7 cm & 13.3 cm \\
Actuator Diameter  & 6.4 cm & 6.7 cm \\
Full Extension Length & 7.6 m & 3 m\\
Maximum Pressure & 10.3 kPa & 10.3 kPa\\
\hline
\end{tabular}
\caption{\textbf{Vine robot parameters for field studies.} Lay-flat diameters, lengths, and maximum pressures of both vine bodies and their respective actuators used for the field studies.}
\label{table:robot parameters}
\end{center}
\end{table}

The robot base, including spool, motor, and encoder; control electronics, including the four closed-loop pressure regulators (QB3, Proportion-Air, McCordsville, IN); and joystick are the same as in~\cite{coad2020vine}. 
The air compressor (Metabo HPT, Braselton, GA) has a maximum flow rate of 113 liters per minute at 620.5~kPa.

\subsection{Initial Field Study Description}

 Fig.~\ref{fig:test_locs} shows the three locations where SPROUT was tested during the initial field study (A), including one entry point at the bottom of the pile accessible to humans (Test~1), as well as two entry points on the top and on the side of the pile, requiring a camera (Insta360 X4) mounted on a pole to monitor SPROUT's behavior (Tests~2 and 3). The robot was run with and without the tip-mounted camera to observe performance. Adding the camera allowed us to meaningfully teleoperate the robot in void spaces where we could not see the robot, but removing the camera allowed us to grow the robot without the added resistance and weight of the camera mount.

\subsection{Initial Field Study Results}

\subsubsection{Test 1: Bottom}
Location 1 (bottom left image of Fig.~\ref{fig:test_locs}) is a 0.66~m wide cavity at the bottom of the pile with an initial horizontal path a human could crawl in. The path then curves right at approximately 90$^{\circ}$, achieving levels 1 and 2 in our taxonomy of difficulty (Table~\ref{table:tod}). Without its camera mount, SPROUT grew into the hole, up and over a 23~cm concrete block, and around the turn (top left image of Fig.~\ref{fig:experiments}) before getting stuck in a corner. SPROUT was unable to grow across a 25~cm gap with a 23~cm increase in height due to pressure leaks that prevented SPROUT from retaining its shape. 



\subsubsection{Test 2: Top}
Location 2 (bottom middle image of Fig.~\ref{fig:test_locs}) was a cavity that could be entered from the top of the pile. The path was initially horizontally straight, but then featured a sharp vertically downward drop, corresponding to difficulty levels 1, 3, and 4. Here, we tested SPROUT without and with its camera mount.  Without the camera mount, SPROUT grew to full length (3~m) at a speed of 0.6~m/min and curved well, supporting itself in free space over a ledge (top middle image of Fig.~\ref{fig:experiments}). With the camera mount, SPROUT got caught on rubble and experienced reduced mobility, but was still able to grow over a horizontal section and over the ledge. When entering the vertical drop section, SPROUT collapsed after bridging approximately a meter into the void space and then struggled to lift its tip from the ground.

\subsubsection{Test 3: Side}
\begin{figure*}[!t]
\centering
\includegraphics[width=\linewidth]{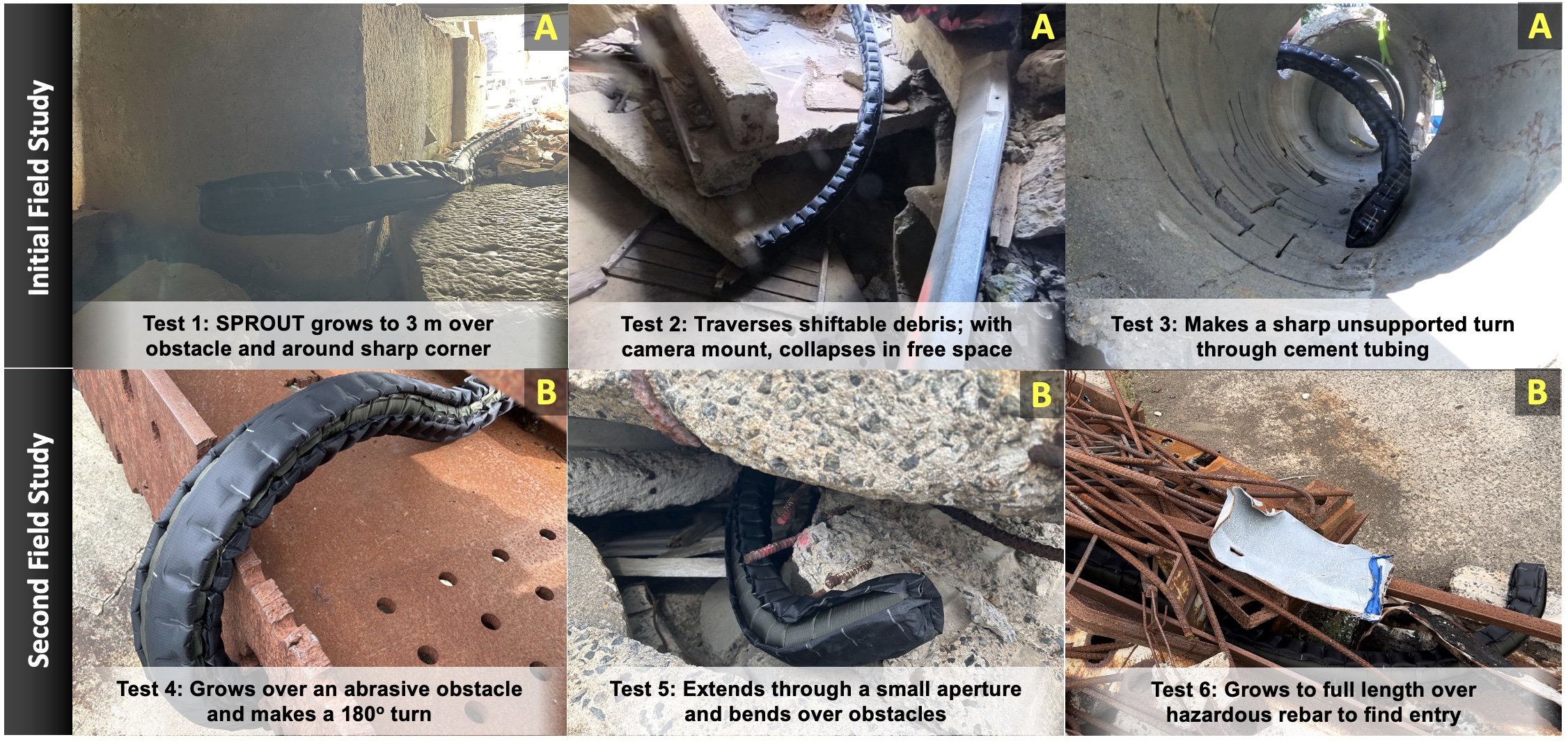}
\caption{\textbf{Key findings from experiments.} Without the camera mount, SPROUT steered in all directions and grew around sharp turns. With the mount, SPROUT had more limited tip mobility. With the hardware improvements between the field studies, SPROUT exhibited more capable maneuverability, faster growth, and more reliable teleoperation.}
\label{fig:experiments}
\vspace{-3ex}
\end{figure*}

Location 3 (bottom right image of Fig.~\ref{fig:test_locs}) was a large pipe on the side of the pile with multiple entry holes, including a small one that was vertical, fulfilling difficulty levels 4 and 5. In this location, without its camera mount, SPROUT completed a sharp turn while unsupported (top right image of Fig.~\ref{fig:experiments}), which allowed the robot to go in the desired direction once inside the pipe.

 \section{POST-IMPROVEMENT FIELD STUDY}

\subsection{Updates for Durability and Portability}
The initial field study highlighted several issues with SPROUT's durability and portability. SPROUT experienced multiple sources of leaks. The first was the connection between the main body tube and the outlet. This seal of high-strength tape melted under \mbox{37$^{\circ}$C} heat and humidity during the initial field study and struggled to maintain a tight coupling at higher pressures. The short heat seals on the robot's actuators also sporadically failed and led to further air loss. We theorized that the 3~mm width of the seals was insufficient, and that the sewing reinforcements intended to reinforce the heat seals actually increased failure rates due to the added thread holes in the actuators. These initial design choices coupled with challenging unfavorable weather conditions resulted in poor durability during repeated pressurization.  

To address these problems, we first changed the seal between the main body tube and the outlet base to two elastic bands and a hose clamp. To eliminate leakages on the actuators, the original actuators were replaced with new ones that had a seal width of 5~mm and no sewing reinforcements. These were found to be more resistant to bursting from cyclic inflation and deflation testing in the lab, with no failures. Finally, we switched the body fabric to a lighter 30 denier TPU-coated ripstop nylon with a thickness of 0.13~mm to further reduce the main body weight. The parameters for this robot are shown in Table~\ref{table:robot parameters}.

Additionally, through our discussions with USAR experts, we determined that the initial SPROUT system would be of limited use to USAR teams since it was not fully portable as a single unit. SPROUT also required a number of independent components, a 120V AC power line, and an air compressor. These power and air sources would be an impractical requirement in an actual disaster scenario. In response to this feedback, we reengineered the base and its associated electrical and compute system to advance the system's overall technological readiness level. Our improvements also prioritized compatibility with existing equipment supplies used by USAR teams. First, instead of using compressed air from a line-powered, noisy air compressor, we switched to a quieter Self-Contained Breathing Apparatus (SCBA) tank. Nominally, this tank contains 30 minutes of air for emergency breathing in HAZMAT scenarios and is a standard part of a USAR equipment loadout. A low-pressure regulator (SCBAS-HPR-1-400, SCBAS, Inc.) was used to step the tank pressure down from 15279 kPa (2216 PSI) to 207 kPa (30 PSI) as expected by the inlet to the pressure regulator for the base. The SCBA tank is carried by the human responder in a backpack harness (Fig.~\ref{fig:parts_image}B), and connected to the main pressure line for SPROUT, and allows for approximately 1 hour of continuous operation. 
\begin{figure*}[!t]
    \vspace{0.5em}
    \centering
    \includegraphics[width=\linewidth]{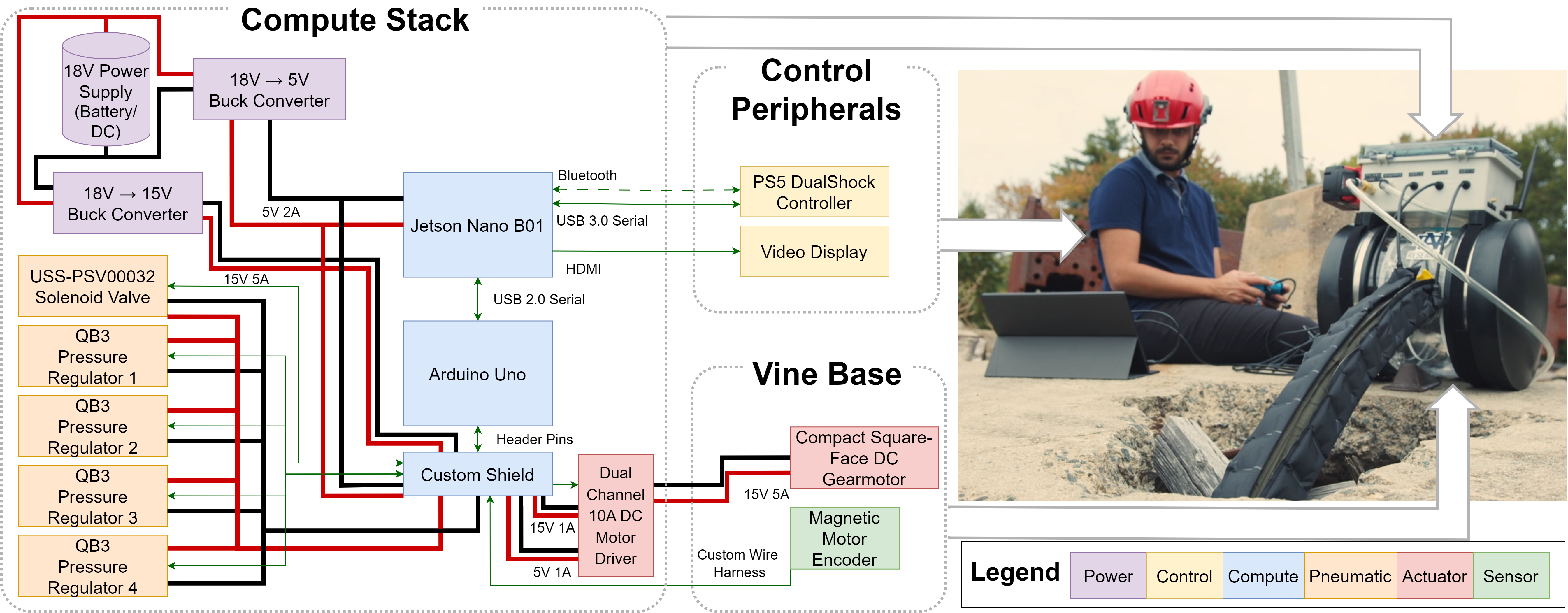}
    \caption{\textbf{System architecture for portable SPROUT operation}. With improvements between field studies, SPROUT runs fully on battery power, features a condensed electronics and fluidic control system, and utilizes a joystick gamepad for control of the vine teleoperation.}
    \label{fig:sys_arch}
    \vspace{-1.5em}
\end{figure*}
To reduce the number of independent components required to operate SPROUT, our updated system includes a compressed electronics printed circuit board (PCB). This PCB is mounted directly on top of the system microcontroller (Arduino Uno Rev3, Arduino.cc) and performs the same function as our original control circuit, which used two solderable breadboards, while saving space.

Instead of an external laptop and joystick used to send commands to the regulators through the microcontroller, a joystick gamepad (DualSense, Play Station) sends operating commands to an NVIDIA Jetson Nano computer running the Robot Operating System (ROS) \cite{Quigley2009}, which interfaces with the microcontroller to set desired pressures on the regulators. 

All components are powered with an 18V, 5.0 Ah tool battery (M18 Red Lithium, Milwaukee), a common item used by USAR responders. The electrical connections between components are detailed in Fig.~\ref{fig:sys_arch}, which describes all components used in SPROUT's updated build. The compute and fluidic actuation systems are contained in an IP67-rated washdown box that sits on top of the SPROUT base, as depicted in Fig.~\ref{fig:parts_image}B. The final robot system has a mass of 13.6 kg, not including the SCBA tank, and is designed to be operated by two-person teams in which one person carries and teleoperates the vine base and the second carries and monitors the compressed air supply, as depicted in Fig.~\ref{fig:parts_image}B.

\subsection{Second Field Study Description}
With the field portability improvements made after the initial field study, we conducted a second field study to determine the new capabilities of SPROUT. Three test locations were chosen, where two of the spaces (Test 4 and Test 6) were spaces that were too hazardous to enter for humans, and two spaces (Test 5 and Test 6) were too small for humans to enter.
For these tests SPROUT was run without the camera mount. SPROUT was able to be continually used for over an hour with just one SCBA tank and a single charged battery. Replacing the SCBA tank and battery takes under three minutes to complete, allowing for at least another hour of operation. 

Throughout the entire duration of the second field study, no actuator seals broke and no leaks on the main body tube occurred. SPROUT was able to be set up faster, be moved to new locations faster, and grow faster (due to less leaks and lubricated guide rod) while exploring a confined space. 

\subsection{Second Field Study Results} \label{sec:second_exp}

\subsubsection{Test 4: Abrasive} \label{abrasive}

Location 4 (top left image of Fig.~\ref{fig:test_locs}) was an unstable, abrasive rusted metal structure that sat on top of the rubble pile. The path included a traversal above a ledge to enter the rusted structure, followed by a flat portion on the straight section of the structure, and then an abrasive edge entry path to a cavity beneath the rusted structure. Without its camera mount, SPROUT successfully overcame these initial obstacles, and grew to its full length of 3~m. SPROUT was then able to inspect the cavity beneath the structure after completing a full 180$^{\circ}$ turn (shown on the bottom left of Fig. \ref{fig:experiments}), growing around an abrasive edge. SPROUT experienced no difficulty in completing this inspection. This test achieved levels 3 and 6 on the taxonomy of difficulty. 

\subsubsection{Test 5: Small}

Location 5 (top middle image of Fig. \ref{fig:test_locs}) was a small entry path with a number of unsupported sections. Without its camera mount, SPROUT entered the cavity through a small aperture, only slightly larger than SPROUT's body size, and then grew over unsupported sections to reach the opening of the cavity on the other side, shown in the bottom middle image of Fig \ref{fig:experiments}. To traverse the path, SPROUT's body was controlled so that it had support along its body at various points, allowing it to fully traverse even the unsupported sections. This test achieved levels 2 and 5 on the taxonomy of difficulty.

\subsubsection{Test 6: Rebar} \label{rebar}
Location 6 (top right image of Fig.~\ref{fig:test_locs}) was a large pile of rusted rebar with a small entry path. Here, without its camera mount, SPROUT grew through the entire section of rebar, extending along sharp and heavily rusted sections with no difficulty, depicted in the bottom right image of Fig. \ref{fig:experiments}. When impeded by crossing sections of rebar, SPROUT was able to use its soft, compliant body to easily grow around the impeding sections and find a path out of the hazardous rebar. This test achieved levels 2, 5 and 6 on the taxonomy of difficulty.

\section{DISCUSSION}
In this section, we discuss the insights gained from both our first and our second field studies, including a comparison of desired capabilities for a USAR search tool with the current capabilities of SPROUT. We also discuss the need-driven challenges that remain to be solved, and what new research opportunities this work has enabled.

\begin{table*}[tb]
\centering
\vspace{0.5em}
\begin{tabular}{c|c}
\hline
\textbf{Desired USAR Capability}  & \textbf{SPROUT Robot} \\ \hline
Long Reach (\(>\)3~m)  & \cellcolor{green!25}Demonstrated to 3 m, longer possible \\
Tortuous Path Traversal & \cellcolor{green!25}Completed multiple paths with turns \\
Tip-Mounted Sensing & \cellcolor{red!25}Collected visual information, but camera mount weight (145~g) restricted steering \\
Fits Through 6.4~cm Drill Hole  & \cellcolor{red!25}Camera mount was 8.5~cm wide and could not compress \\
Remote Teleoperation  & \cellcolor{green!25}Demonstrated 8.4~m between human operator and robot base via wired connection \\
Intuitive Control  & \cellcolor{red!25}Situational awareness challenging from tip camera image alone\\
Portability & \cellcolor{green!25}Portable with a two-person team \\
Durability & \cellcolor{green!25}Robot lasted through second field study, no damage from environment\\
Speed & \cellcolor{red!25}2.5 m/min growth, slower with camera mount \\
\hline
\end{tabular}%
\caption{\textbf{Field performance of SPROUT system}. Desired capabilities for USAR are compared to those realized on the SPROUT system. Color Key: Green, success; Red, improvement needed.}
\label{table:challenges}
\vspace{-2.0em}
\end{table*}

\subsection{Field Study Insights}

Based on our discussions with USAR experts and our experience testing SPROUT in the field, we have created Table~\ref{table:challenges}, which summarizes
the desired capabilities for a robotic search tool for use in USAR, compared to the current capabilities of SPROUT. We are primarily testing SPROUT against existing USAR search and inspection cameras, which have limited extension and minimal steering.

Ultimately, SPROUT performed well without its camera mount, growing to a length of 3~m and steering in all directions in environments both easy and challenging for humans. 
With the camera mount, SPROUT sometimes struggled to grow due to dimensional tolerance issues: the inner diameter of the rigid tip mount was too small and restricted the robot from everting its actuators while steering. Also, as previously noted in \cite{coad2020vine}, the mount's rigidity and location outside the robot body prevented the robot from squeezing through an aperture smaller than the mount and caused it to drag on the environment. The mount's added mass (145~g) also led to a restricted workspace when steering. We aim to redesign the camera mount to be more useful in USAR situations. 

SPROUT was successfully able to be operated remotely via a joystick over an 8.4~m cable. However, the control of the robot based on the tip camera image alone was not intuitive, in part due to the limited field of view of the camera and the robot's slow growth, and also because the robot tip tended to rotate when it steered, causing the camera image to not align with gravity down. 
SPROUT's growth speed should also be increased, as it currently grows at 2.5~m/min without its camera mount and slower with the mount. For USAR, response speed is critical, so the faster the robot can grow, the more useful it will likely be.



\subsection{Challenges}

After these field studies, it is necessary to analyze some of the challenges SPROUT still faces that may impede its use for USAR. A primary challenge is carrying a load with the robot body. Vine robots are strongest in tension, and adding weight, especially to the tip, can severely limit its ability to lift and steer its tip, as well as cross gaps in the environment. Increasing the internal body pressure would allow the SPROUT system to support more load, ideally making it possible to carry more sensors into the environment. However, increasing pressure makes bursting and leaks more likely and achieving high curvature steering more difficult.

An additional challenge to carrying sensors into the environment is that, while the vine body itself is compliant, the sensors and their associated mounts are rigid. As we saw with the camera mount, these rigid components can provide serious limitations on the spaces SPROUT can travel into, especially when entering a 6.4~cm drill hole. Externally mounted sensors can also become caught on the environment. In a worst case scenario, the robot could be abandoned in the inspection space and be easily replaced since the vine body itself is inexpensive. However, the sensors on the tip of the vine are potentially more costly.


Vine robots also must explore potentially hazardous environments and hence must be durable to a variety of dangers. While we made many improvements to the durability of the SPROUT system, these changes largely focused on durability to the robot's pressurization. The act of eversion allows the robot to grow without sliding relative to the ground. However, when the robot steers, it can scrape against the ground, along with any external mounts it is carrying.  Aside from the ground, USAR situations are unpredictable and an environment could contain water, electrical hazards, and unknown debris, all of which could damage the robot or its sensors. The robot could grow after being punctured, but would have reduced capabilities, especially for bearing load. Therefore, we must work to make the system robust against the many obstacles a USAR situation can contain. In the results presented in sections \ref{abrasive} and \ref{rebar}, we showed that the SPROUT system is robust against abrasive edges and rebar, but testing on other hazards has yet to be completed.

\subsection{Opportunities}

Despite these challenges, there are also many opportunities for growth with the SPROUT system. While our demonstrated robots were only able to grow to 3~m, vine robots have been demonstrated to grow as long as 70~m. Their maximum length is largely restricted by friction of the body material with itself, and the ability to store all the material within a base unit. While 70~m is likely much longer than would ever be needed for a USAR scenario, we have shown that a growing robot can function in these spaces. With this ability to grow, vine robots can easily surpass the length of current telescoping search cameras.

Vine robots also have a greater ability to steer than telescoping cameras or push cameras since they can complete multiple turns. They can conform to the environment, steer their entire body, and articulate just their tip with certain steering schemes. While we were not able to demonstrate SPROUT's steering at length, we did show it was capable of steering within these complicated spaces. 

Ultimately, vine robots have a form factor that allows for entering small spaces that humans cannot access. Provided the mount is kept small, a vine robot can squeeze through tight apertures, possibly even smaller than the 6.4~cm drill hole needed for a traditional camera. Vine robots can grow through existing gaps in the environment as well. By increasing SPROUT's maximum pressure range, we can advance its ability to push through these gaps and advance through these spaces faster. This goal of fitting through smaller holes also incentivizes work to make actuators lower profile and sensors for the system more compact.

\section{CONCLUSIONS}
In this work, we presented the results from two separate field studies of the SPROUT system at the MA-TF1 training site. Our initial field study included three tests demonstrating the robot's ability to steer and use its camera in these spaces. However, it also highlighted the need for improved durability and portability. After making design changes, we conducted our second field study, which showed the robot's improved ability to grow and steer in challenging spaces. With these tests, it is clear that the SPROUT system is highly capable of navigating this class of sites. 


In future studies, we plan to mount additional sensors on the tip of the vine, such as IMUs and depth cameras. Additionally, we will experiment with smaller, lighter, and compliant camera mount designs that do not impede the robot's movement while still fitting through small spaces. We also plan to improve the user's situational awareness by rotating the camera view as the vine body contorts when exploring void spaces. To improve growth speed, we will experiment with improving the robot base's ability to hold pressure, and use lower-friction materials to fabricate the vine body. Finally, we plan to map SPROUT's complete workspace to better model its movement within these spaces. This will enable more reliable growth and path planning, reducing the need to retract and start over after an error. Overall, our work demonstrates that vine robots are platforms with great potential to affect USAR operations, especially when developments are tightly coupled to the operating conditions of the field.





\section*{ACKNOWLEDGMENTS}

Thanks to members of Massachusetts Task Force 1 for providing access to their training site and feedback on the robot's capabilities. Thanks to Kevin Wakakuwa, Lydia Chau, Allison Fick, and Myia Dickens for assisting with fabrication of the robot. Thanks to Adam Norige and John Aldridge for offering insightful comments in the framing of this work.


\bibliographystyle{IEEEtran} 
\bibliography{references}

\addtolength{\textheight}{-12cm}   

\end{document}